\pdfoutput=1

\documentclass[11pt]{article}

\usepackage[preprint]{acl}

\usepackage{times}
\usepackage{latexsym}

\usepackage[T1]{fontenc}

\usepackage[utf8]{inputenc}

\usepackage{microtype}

\usepackage{inconsolata}

\usepackage{graphicx}

\usepackage{amsmath,amsfonts,bm}

\def\eqref#1{equation~\ref{#1}}

\def\1{\bm{1}}

\DeclareMathAlphabet{\mathsfit}{\encodingdefault}{\sfdefault}{m}{sl}
\SetMathAlphabet{\mathsfit}{bold}{\encodingdefault}{\sfdefault}{bx}{n}

\usepackage{colortbl}
\usepackage{hyperref}
\usepackage{url}
\usepackage[linesnumbered,ruled,vlined]{algorithm2e}
\usepackage{amsmath}
\usepackage{multirow}
\usepackage{tabularx}
\usepackage{float}
\usepackage{subfigure} 
\usepackage{algpseudocode}
\usepackage{multirow}
\usepackage{array}
\usepackage{siunitx}
\usepackage{booktabs}
\usepackage{hyperref}
\usepackage{enumitem}
\usepackage{balance}

\usepackage{appendix}
\usepackage{titletoc}
\usepackage{wrapfig}
\usepackage{listings}
\usepackage{xcolor}
\newcommand{\hide}[1]{} 

\newcommand{\vpara}[1]{\vspace{1.5ex}\noindent\textbf{#1}}

\newcommand{\model}{LoRS}
\newcommand{\smodel}{LoRS }

\title{LoRS: Efficient Low-Rank Adaptation for Sparse Large Language Model}

\author{
    Yuxuan Hu\textsuperscript{\rm 1,\rm 2},\ Jing Zhang\textsuperscript{\rm 1,\rm 2}\thanks{Corresponding author. zhang-jing@ruc.edu.cn},\ Xiaodong Chen\textsuperscript{\rm 1,\rm 2}\\
    {\bf Zhe Zhao}\textsuperscript{\rm 4},\ {\bf Cuiping Li}\textsuperscript{\rm 1,\rm 3},\ {\bf Hong Chen}\textsuperscript{\rm 1,\rm 3} \\
    \textsuperscript{\rm 1}School of Information, Renmin University of China, Beijing, China \\
    \textsuperscript{\rm 2}Key Laboratory of Data Engineering and Knowledge Engineering, Beijing, China \\
    \textsuperscript{\rm 3}Engineering Research Center of Database and Business Intelligence, Beijing, China \\
    \textsuperscript{\rm 4}Tencent AI Lab, Beijing, China \\
}

\begin{document}
\maketitle

\begin{abstract}
Existing low-rank adaptation (LoRA)  methods face challenges on sparse large language models (LLMs) due to the inability to maintain sparsity. Recent works introduced methods that maintain sparsity by augmenting LoRA techniques with additional masking mechanisms. Despite these successes, such approaches suffer from an increased memory and computation overhead, which affects efficiency of LoRA methods. In response to this limitation, we introduce LoRS, an innovative method designed to achieve both memory and computation efficiency when fine-tuning sparse LLMs. To mitigate the substantial memory and computation demands associated with preserving sparsity, our approach incorporates strategies of weight recompute and computational graph rearrangement. In addition, we also improve the effectiveness of LoRS through better adapter initialization. These innovations lead to a notable reduction in memory and computation consumption during the fine-tuning phase, all while achieving performance levels that outperform existing LoRA approaches.
\end{abstract}


\section{Introduction}

Large language models (LLMs)~\cite{llama2-touvron2023, llama3-dubey2024} have demonstrated remarkable proficiency in numerous natural language processing tasks, which has spurred their increasing integration into diverse applications. However, the deployment of these models is constrained by their vast parameter counts, necessitating significant hardware resources that can be prohibitive for many users. Moreover, the large scale of LLMs can impede inference speed, presenting a challenge in scenarios requiring rapid response times.

To mitigate these issues, various post-training pruning methods have been introduced, such as SparseGPT~\cite{sparsegpt-frantar2023}, Wanda~\cite{wanda-sun2024}, and RIA~\cite{RIA-zhang2024}. These techniques effectively reduce model parameters, transforming dense models into sparse versions with minimal data requirements and within short periods. Despite their efficiency, pruned models still exhibit a performance disparity compared to their original counterparts, especially in small and medium-sized models with unstructured or 2:4 semi-structured sparsity~\cite{2:4sparsity-mishra2021}. This discrepancy limits the practical utility of pruned models. Continuous pre-training could help bridge this gap but comes at a high computational cost. Consequently, there is a pressing need for tuning methods that maintain sparsity while optimizing memory and parameter efficiency.

Low-Rank Adaptation (LoRA)~\cite{lora-hu2021} was developed to ease the computational demands of training dense LLMs. LoRA enables fine-tuning with reduced resource consumption, making it widely applicable for dense models. Recent studies SPP~\cite{spp-lu2024} and SQFT~\cite{sqft-munoz-etal-2024}, have extended LoRA to accommodate sparse LLMs by incorporating masking mechanisms. We refer to these methods as Sparsity Preserved LoRA methods (SP-LoRA). SPP and SQFT achieving performance similar to LoRA, while ensuring the sparsity of the model. However, they increase computation and memory overhead, undermining LoRA's inherent efficiency. Specifically, SQFT requires twice the memory overhead of LoRA, while SPP reduces the memory overhead to the same as LoRA through gradient checkpoints~\cite{gradckpt-chen2016}, but greatly increases the time overhead.

\begin{figure*}[t]
  \centering
  \includegraphics[width=0.95\textwidth]{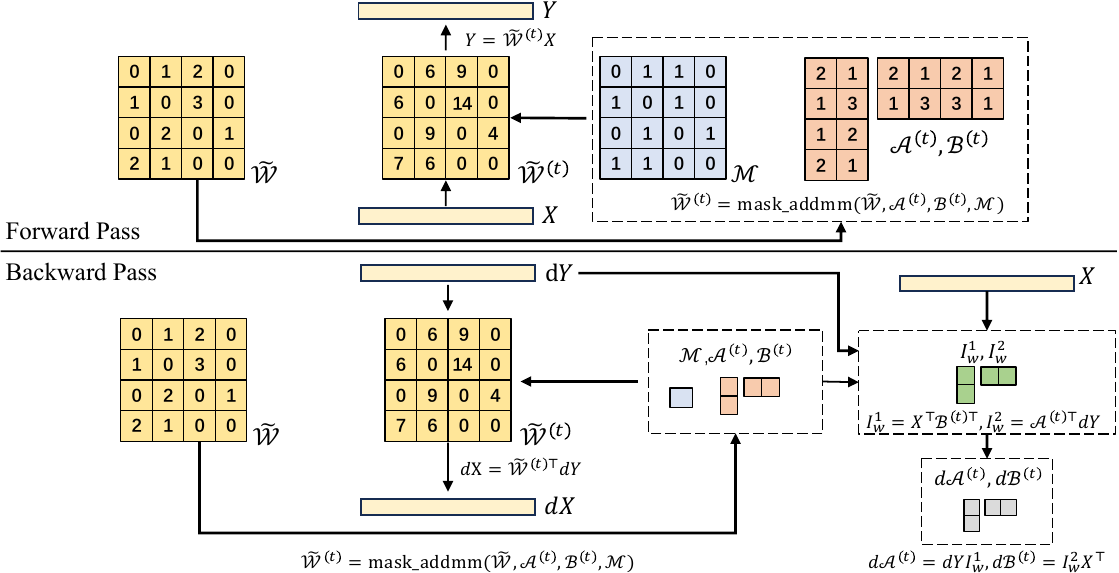}
  \caption{The workflow of \model.\label{Figure: workflow}}
\end{figure*}

In response to these limitations, we present an innovative \textbf{Lo}w \textbf{R}ank Adaptation method for \textbf{S}parse LLM (LoRS). LoRS addresses the increased memory and computational overhead caused by masking mechanisms through weight recompute, and computational graph rearrangement. Our approach discards the fitness weights during each forward pass and recalculates them during backward passes, thereby significantly reducing the memory overhead at the cost of a small amount of additional computation. Meanwhile, we optimize the gradient computation by computation graph rearrangement in the backward pass, which further reduces the computational overhead compared to SQFT and SPP. In addition, inspired by the latest LoRA variants, we also improve the efficiency of LoRS by better adapter initialization.

We evaluate LoRS on multiple LLMs, initially pruning them via post-training methods like Wanda or SparseGPT. Subsequently, LoRS is used to fine-tune these models using instruction datasets or pretraining datasets. The zero-shot performance of the tuned sparse LLMs is then assessed across a variety of benchmark tasks. The main contributions of this paper are summarized in the following:

(1) We introduce LoRS, a novel fine-tuning method for sparse LLMs that preserves sparsity while minimizing computation and memory overhead. LoRS leverages weight recompute and computational graph rearrangement techniques to achieve this efficiency and achieve better performance through better adapter initialization.

(2) Through comprehensive experiments on sparse LLMs with different sparsity patterns, we show that LoRS can outperform existing SP-LoRA methods in terms of performance, memory usage, and computation efficiency.

\section{LoRS}
In this section, we begin by reviewing unstructured pruning and low-rank adaptation in Section~\ref{sec: preliminary}. We then proceed to analyze the memory complexity associated with existing methods in Section~\ref{sec: complexity analysis}. We then describe how our method LoRS optimizes the memory and computational overhead of existing methods in section ~\ref{sec: memory optimization}. Finally, in Section~\ref{sec: performance optimization}, we describe how the performance of LoRS can be improved by better adapter initialization.

\subsection{Preliminary\label{sec: preliminary}}

\vpara{Unstructured Pruning.} Unstructured pruning~\cite{sparsegpt-frantar2023, wanda-sun2024, RIA-zhang2024} converts dense weight matrices of LLMs into sparse matrices to enhance computational efficiency. Given the original dense weight matrix $\mathcal{W} \in \mathbb{R}^{R \times C}$, pruning aims to produce a sparse matrix $\tilde{\mathcal{W}}$ through the application of a binary mask $\mathcal{M} \in \{0, 1\}^{R \times C}$ and weight updates $\Delta \mathcal{W} \in \mathbb{R}^{R \times C}$. This process is mathematically represented as: $\tilde{\mathcal{W}} = \mathcal{M} \odot (\mathcal{W} + \Delta \mathcal{W})$, where $\odot$ denotes element-wise multiplication. The mask $\mathcal{M}$ zeros out less important weights, while $\Delta \mathcal{W}$ fine-tunes the retained weights, ensuring that the pruned model preserves its performance. 


\vpara{LoRA.} Low-Rank Adaptation~\cite{lora-hu2021, loraga-wang2024} is an efficient approach designed to fine-tune LLMs for specific tasks or domains by training only a limited set of parameters. This method allows the model to be adapted to specific tasks while significantly reducing computational cost.

The mathematical representation of LoRA is expressed as $\mathcal{W}^{(t)} = \mathcal{W} + \mathcal{A}^{(t)} \times \mathcal{B}^{(t)}$, where $\mathcal{W}$ stands for the initial weight matrix of the pre-trained model. The term $\mathcal{W}^{(t)}$ denotes the adapted weight matrix at the $t$-th iteration of training. The matrices $\mathcal{A}^{(t)}$ and $\mathcal{B}^{(t)}$ represent the trainable adapter matrices at the $t$-th iteration. Specifically, $\mathcal{A} \in \mathbb{R}^{R \times r}$ and $\mathcal{B} \in \mathbb{R}^{r \times C}$, with $r$ being much smaller in dimension compared to $R$ and $C$. Here, $R$ and $C$ represents the dimensions of the original weight matrix. In practice, during the adaptation process, only the parameters within $\mathcal{A}$ and $\mathcal{B}$ are updated, while all other parameters remain fixed. This strategy ensures that the model can be efficiently tuned to new tasks or domains without altering the entire pre-trained weights.

\vpara{SP-LoRA.} To maintain the sparsity of the model while adaptation, SP-LoRA methods~\cite{spp-lu2024, sqft-munoz-etal-2024} integrate a masking mechanism within the LoRA framework. Let us consider a sparse large language model (LLM) with a weight matrix $\tilde{\mathcal{W}}$ and its associated mask $\mathcal{M}$. During each training iteration $t$, the mask is applied to enforce the sparsity of the weight matrix, which can be mathematically represented as:
\begin{equation}
\tilde{\mathcal{W}}^{(t)} = \tilde{\mathcal{W}} + \mathcal{A}^{(t)} \times \mathcal{B}^{(t)} \odot \mathcal{M} .
\end{equation}

Here, $\odot$ denotes element-wise multiplication, while $\mathcal{A}^{(t)}$ and $\mathcal{B}^{(t)}$ represent adapter matrices that are updated at each iteration.

The incorporation of the mask, while ensuring that the weights remain sparse, modifies the computational graph of the original LoRA framework. This modification results in increased GPU memory usage and computation overhead, presenting practical challenges. Therefore, we will first investigate the reasons behind this elevated GPU memory and computation consumption, and subsequently propose an effective solution to mitigate this issue.

\begin{algorithm}[t]
\SetAlgoLined
\KwIn{Activation $X$, Sparse weight matrix $\tilde{\mathcal{W}}$, \smodel adapters $\mathcal{A}^{(t)}, \mathcal{B}^{(t)}$.}
\KwOut{Activation $Y$}
\BlankLine
Update $\tilde{\mathcal{W}}$ to $\tilde{\mathcal{W}}^{(t)}$: $\tilde{\mathcal{W}}^{(t)} = \tilde{\mathcal{W}} + \mathcal{A}^{(t)} * \mathcal{B}^{(t)} \odot (\tilde{\mathcal{W}} \neq 0)$\;
Save $X$ into context for backward\;
Compute $Y$: $Y = \tilde{\mathcal{W}}^{(t)} X$\;
\caption{LoRS Forward Pass\label{algo: LoRS Forward Pass}}
\end{algorithm}

\begin{algorithm}[t]
\SetAlgoLined
\KwIn{Gradient $dY$, Activation $X$, Sparse weight matrix $\tilde{\mathcal{W}}^{(t)}$, \smodel adapters $\mathcal{A}^{(t)}, \mathcal{B}^{(t)}$.}
\KwOut{Gradients $d\mathcal{A}^{(t)}$, $d\mathcal{B}^{(t)}$, and $dX$}
\BlankLine
Recompute weight $\tilde{\mathcal{W}}^{(t)}$: 
$\tilde{\mathcal{W}}^{(t)} = \tilde{\mathcal{W}} + \mathcal{A}^{(t)} * \mathcal{B}^{(t)} \odot (\tilde{\mathcal{W}} \neq 0)$\;
Compute gradient of $X$: 
$dX =\tilde{\mathcal{W}}^{(t)\top} dY$\;
Compute intermediate weight $I_{w}^{1}$: 
$I_{w}^{1} = X^\top \mathcal{B}^{(t)\top}$\;
Compute intermediate weight $I_{w}^{2}$: 
$I_{w}^{2} = \mathcal{A}^{(t)\top} dY$\;
Compute gradient of $\mathcal{A}^{(t)}$: 
$d\mathcal{A}^{(t)} = dY I_{w}^{1}$\;
Compute gradient of $\mathcal{B}^{(t)}$: 
$d\mathcal{B}^{(t)} = I_{w}^{2} X^\top$\;
\caption{LoRS Backward Pass\label{algo: LoRS Backward Pass}}
\end{algorithm}

\begin{figure*}[t]
  \centering
    \begin{minipage}[t]{0.45\textwidth}
      \includegraphics[width=\textwidth]{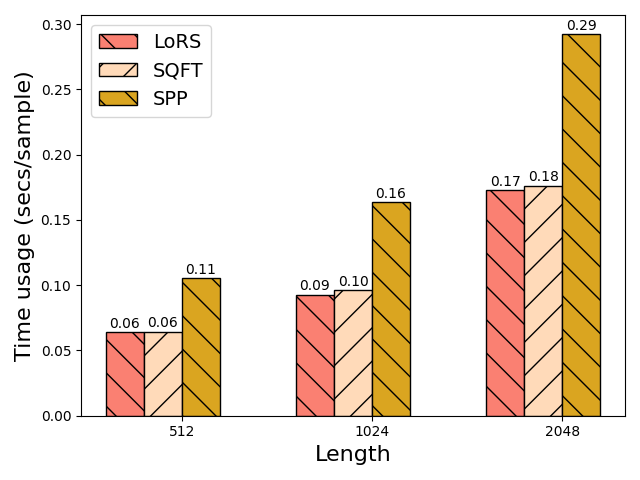}
      \caption{Time usage of LoRA, LoRS, SQFT and SPP.\label{Figure: time usage}}    
    \end{minipage}
    \begin{minipage}[t]{0.45\textwidth}
      \includegraphics[width=\textwidth]{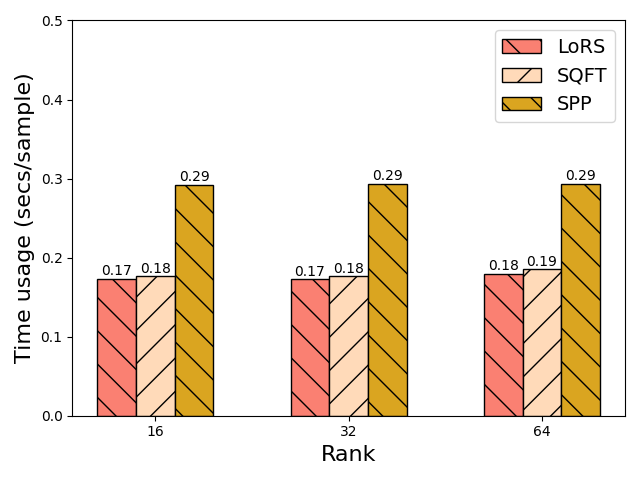}
      \caption{Time usage of LoRS, SQFT and SPP.\label{Figure: time usage rank}}
    \end{minipage}
\end{figure*}

\begin{figure}[t]
  \centering
  \includegraphics[width=0.9\linewidth]{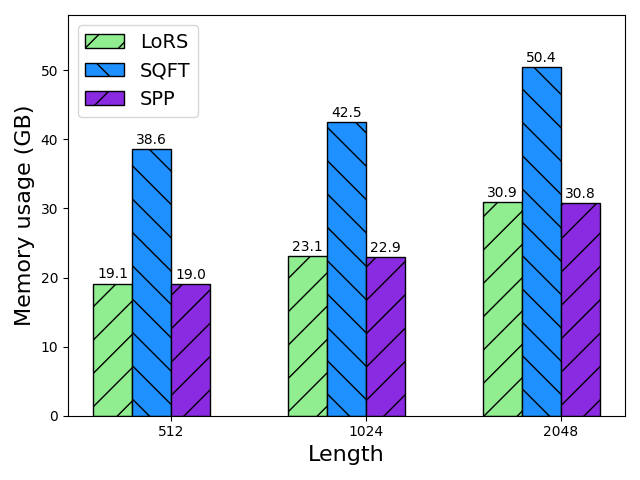}
  \caption{Memory usage of LoRS, SQFT and SPP.\label{Figure: memory usage}}
\end{figure}

\subsection{Complexity Analysis\label{sec: complexity analysis}}

At the \(t\)-th training iteration, let us denote the input to the weight matrix as \(X \in \mathbb{R}^{C \times L}\). For LoRA, the output can be mathematically represented as

\begin{equation}
    Y = \tilde{\mathcal{W}} X + \mathcal{A}^{(t)} \mathcal{B}^{(t)} X.
\end{equation}

The computation process unfolds in these steps:

\begin{equation*}
\begin{array}{ll}
I_{a}^{1} = \tilde{\mathcal{W}} X,       & I_{a}^{2} = \mathcal{B}^{(t)} X, \\
I_{a}^{3} = \mathcal{A}^{(t)} I_{a}^{2}, & Y = I_{a}^{1} + I_{a}^{3},
\end{array}    
\end{equation*}

\noindent where \(I_{a}^{1}\), \(I_{a}^{2}\), and \(I_{a}^{3}\) represent intermediate activations with dimensions \(R \times L\), \(r \times L\), and \(R \times L\) respectively. During back-propagation, gradients for \(\mathcal{A}^{(t)}\), \(\mathcal{B}^{(t)}\), and \(X\) are computed based on the gradient of \(Y\), denoted as \(dY\). The gradient computations are formulated as follows:

\begin{equation*}
\begin{array}{ll}
d\mathcal{A}^{(t)} = dY I_{a}^{2\top}, & I_{a}^{4} = \mathcal{A}^{(t)\top} dY, \\
d\mathcal{B}^{(t)} = I_{a}^{4} X^\top, & I_{a}^{5} = \tilde{\mathcal{W}}^\top dY, \\
I_{a}^{6} = \mathcal{B}^{(t)\top} I_{a}^{4}, & dX = I_{a}^{5} + I_{a}^{6}.
\end{array}   
\end{equation*}

In the forward pass, the input \(X\) and intermediate activation \(I_{a}^{2}\) are stored for back-propagation, involving \(rL + CL\) parameters. Meanwhile, the corresponding multiply–accumulate operations (MACs) for forward is \(RCL+rCL+rRL\) and for backward is \(RCL+2rRL+2rCL\).


For SP-LoRA, the output expression modifies to

\begin{equation}
    Y = (\tilde{\mathcal{W}} + \mathcal{A}^{(t)} \times \mathcal{B}^{(t)} \odot \mathcal{M}) X,
\end{equation}

\noindent where \(\mathcal{M}\) acts as a mask indicating non-zero elements in \(\tilde{\mathcal{W}}\). Unlike LoRA, SP-LoRA requires computing \(\mathcal{M} \odot (\mathcal{A}^{(t)} \times \mathcal{B}^{(t)})\) before multiplying by \(X\). This sequence of operations is outlined as:

\begin{equation*}
\begin{array}{ll}
I_{w}^{1} = \mathcal{A}^{(t)} \mathcal{B}^{(t)}, & I_{w}^{2} = \mathcal{M} \odot I_{w}^{1}, \\
I_{w}^{3} = \tilde{\mathcal{W}} + I_{w}^{2},     & Y = I_{w}^{3} X.
\end{array}    
\end{equation*}

\noindent Meanwhile, back-propagation for SP-LoRA involves:

\begin{equation*}
\begin{array}{ll}
dX = I_{w}^{3\top} dY, & I_{w}^{4} = dY X^\top, \\
I_{w}^{5} = I_{w}^{4} \odot \mathcal{M}, & d\mathcal{A}^{(t)} = I_{w}^{5} \mathcal{B}^{(t)\top}, \\
d\mathcal{B}^{(t)} = \mathcal{A}^{(t)\top} I_{w}^{5}. &
\end{array}    
\end{equation*}

\noindent SP-LoRA's forward pass necessitates retaining \(X\), \(\mathcal{M}\), and \(I_{w}^{3}\) for back-propagation including \(2RC + CL\) parameters, and the MACs corresponding to the forward and backward are \(RCL+RC+rRC\) and \(2RCL+2rRC+RC\), respectively.


Based on frequently used model sizes and training configurations, we assume that \(r \ll R \approx C \approx L\). Comparing LoRA and SP-LoRA, it can be seen that incorporating masks in SP-LoRA significantly raises GPU memory overhead due to traced mask \(\mathcal{M}\) and weight matrix \(I_w^3\) in computational graph (\(rL + CL \to 2RC + CL \Rightarrow \approx 2RC \uparrow\)). Meanwhile, SP-LoRA requires additional computation in the backward pass due to the need to compute the gradient of the weight matrix (\(RCL+2rRL+2rCL \to 2RCL+2rRC+RC \Rightarrow \approx RCL \uparrow\)). Therefore, optimizing GPU memory usage and computation overhead in SP-LoRA is essential.

In this work, we consider the most advanced SP-LoRA methods SPP and SQFT, where SQFT does not take into account the memory and computational overheads and has the same complexity as analyzed above. SPP, on the other hand, optimizes memory usage through PyTorch's built-in gradient checkpoint API, and its implementation is shown in Appendix~\ref{Appendix: Implementation}. Thus, on top of LoRA, SPP reduces the number of parameters that need to be stored in the computational graph (\(rL + CL \to CL \Rightarrow \approx rL \downarrow\)), but introduces additional computation during backward pass (\(RCL+2rRL+2rCL \to 3RCL+3rRC+2RC \Rightarrow \approx 2RCL \uparrow\)).

\subsection{Memory and Computation Optimization}
\label{sec: memory optimization}
After comparing the computational processes of LoRA and SP-LoRA, it is evident that the memory overhead in SP-LoRA arises from the need to maintain additional masks and adapted weight matrices within the computational graph, and the computation overhead arises from the need to compute the gradient of weight matrices. 


To address the memory overhead, inspired by gradient checkpoint~\cite{gradckpt-chen2016}, we introduce weight recompute strategie in LoRS, effectively eliminating the necessity for masks and adapted weight matrices in the computation graph. Specifically, we release the intermediate weights \(\mathcal{M}\) and \(I_{w}^{3}\) directly after the forward pass of the LoRS, and recompute them later during the backward pass. After optimization, only the input activation \(X\) is saved to the computational graph for subsequent backward pass. With this optimization, for each linear layer, we reduce the recorded parameter from \(2RC + CL\) to \(CL\), while only increasing the computational overhead of \(rRC\) MACs.

After that, to reduce the computation overhead associated with computing the gradient of the weight matrix, we propose the computational graph reordering method. Firstly, we find that the masking operation of the gradient during backward pass (\(I_{w}^{5} = I_{w}^{4} \odot \mathcal{M}\)) has minimal effect on the model performance and thus can be ignored, which is equivalent to estimating the gradient using straight through estimator~\cite{STE-bengio2013, SR-STE-zhou2021}. After that, we can directly compute gradients \(d\mathcal{A}^{(t)}\) and \(d\mathcal{B}^{(t)}\) based on \(\mathcal{A}^{(t)}\), \(\mathcal{B}^{(t)}\), \(X\), and \(dY\), i.e.,
\begin{align*}
    & d\mathcal{A}^{(t)} = dY X^\top \mathcal{B}^{(t)\top}, \\
    & d\mathcal{B}^{(t)} = \mathcal{A}^{(t)\top} dY X^\top.
\end{align*}

\noindent Instead of following the computational graph and prioritizing the computation of \(dY X^T\), we can reorder the computation process to compute \(X^\top \mathcal{B}^{(t)\top}\) and \(\mathcal{A}^{(t)\top} dY\) first, thus reducing the MACs from \(RCL + 2rRC\) to \(2rCL + 2rRL\). The optimized backward propagation processes are as follows:


\begin{equation*}
\begin{array}{ll}
dX = \tilde{W}^{t\top} dY, & I_{w}^{1} = X^\top \mathcal{B}^{(t)\top}, \\
I_{w}^{2} = \mathcal{A}^{(t)\top} dY, & d\mathcal{A}^{(t)} = dY I_{w}^{1}, \\
d\mathcal{B}^{(t)} = I_{w}^{2} X^\top.&
\end{array}    
\end{equation*}


Finally, the workflow of LoRS is illustrated in Figure~\ref{Figure: workflow}. Meanwhile, algorithm~\ref{algo: LoRS Forward Pass} and~\ref{algo: LoRS Backward Pass} details the forward pass and backward pass of LoRS. It can be seen that after optimization, LoRS only needs to store \(rL\) parameters in the computational graph, and at the same time, it only needs the MACs of \(RCL + rRC + RC\) in the forward pass and \(RCL + 2rRL + 2rCL + rRC + RC\) in the backward pass. Compared to LoRA, LoRS reduces the parameters stored in the computational graph and increases only the MACs of \(rRC + RC\), superior to SP-LoRA, which increases the \(2RC\) parameters stored in the computational graph and increases the MACs of \(RCL\).


\subsection{Performance Optimization}
\label{sec: performance optimization}

The existing SP-LoRA methods SPP and SQFT use zero initialization and random initialization to initialize the adapters. However, recent advances in LoRA variants highlight the critical impact of initialization strategies on overall performance. Drawing inspiration from the methodologies of LoRA-GA~\cite{loraga-wang2024}, we introduce a gradient-based initialization technique aimed at enhancing LoRS's effectiveness.

Referring to existing LoRA variants, we initialize \(\mathcal{A}^{(0)}\) to 0, while minimizing the difference between LoRS and full fine-tuning on the first training iteration by initializing \(\mathcal{B}^{(0)}\). To illustrate, with \(A^{(0)} = 0\) and disregarding masking operations, we derive the following equations:
\begin{align*}
    & dA^{(0)} = d\tilde{W}^{(0)} B^{(0)\top}, dB^{(0)} = A^{(0)\top}d\tilde{W}^{(0)}, \\
    & \tilde{W}^{(1)} - \tilde{W}^{(0)} = A^{(1)} B^{(1)}\\
    & \quad = (A^{(0)} + dA^{(0)}) (B^{(0)} + dB^{(0)}) \\
    & \quad = dA^{(0)}B^{(0)} \\
    & \quad = d\tilde{W}^{(0)}B^{(0)\top}B^{(0)}.
\end{align*}

\noindent This derivation illustrates the relation between the adapters and the gradient obtained from the initial training step. Therefore, we determine \(B^{(0)}\) through an optimization process as follow:
\begin{equation}
    B^{(0)} = \underset{B}{\arg\min} \Vert d\tilde{W}^{(0)} - d\tilde{W}^{(0)} B^\top B \Vert
\end{equation}

\noindent This optimization objective can be solved by singular value decomposition, i.e., \(U, S, V = {\rm SVD}(d\tilde{W}^{(0)}), B^{(0)} = V_{:r}\).

While the gradient-based initialization does require access to the gradients of the weight matrices during the first training step, we adopt layer-by-layer initialization strategy to ensure that this process can be carried out without imposing additional memory costs. This efficient initialization paves the way for improved performance in subsequent training iterations.
\section{Experiments}

\newcommand{\B}[1]{\textbf{#1}}




\begin{table*}[t]
\newcolumntype{?}{!{\vrule width 1pt}}
\newcolumntype{C}{>{\centering\arraybackslash}p{2em}}
\centering

\renewcommand\arraystretch{1.5}

\resizebox{\linewidth}{!}{
\begin{tabular}{cllcccccccc}
\hline
Model                        & Method            &  Sparsity & ARC-c & ARC-e & BoolQ & Hellaswag & OBQA  & RTE   & Winogrande & Average \\ \hline
Llama-2-7B                   & None              &  None     & 43.52 & 76.35 & 77.74 & 57.14     & 31.40 & 62.82 & 69.06      & 59.72   \\ \cline{2-11} 
                             & SparseGPT         &  2:4      & 31.31 & 63.93 & 68.90 & 43.54     & 24.60 & 63.18 & 65.90      & 51.62   \\
                             & SparseGPT+SPP     &  2:4      & 36.86 & 69.15 & 72.91 & 50.67     & 28.80 & 62.45 & 66.30      & 55.31   \\
                             & SparseGPT+SQFT    &  2:4      & 36.01 & 64.35 & 72.17 & 51.84     & 29.60 & 59.93 & 63.61      & 53.93   \\
                             \rowcolor{gray!10}
\cellcolor{white}            & SparseGPT+LoRS    &  2:4      & \B{37.63} & \B{70.03} & \B{74.22} & \B{51.95}     & \B{30.20} & \B{63.90} & \B{66.38}      & \B{56.33}   \\ \cline{2-11} 
                             & Wanda             &  2:4      & 30.03 & 61.95 & 68.32 & 41.21     & 24.20 & 53.07 & 62.35      & 48.73   \\
                             & Wanda+SPP         &  2:4      & 36.26 & 69.44 & 72.02 & 49.64     & 27.80 & 55.96 & 63.77      & 53.56   \\
                             & Wanda+SQFT        &  2:4      & 35.41 & 65.03 & \B{72.39} & 50.18     & \B{30.00} & \B{60.29} & 62.67      & 53.71   \\ 
                             \rowcolor{gray!10}
\cellcolor{white}            & Wanda+LoRS        &  2:4      & \B{37.12} & \B{70.71} & 71.56 & \B{51.18}     & 27.60 & 57.76 & \B{64.48}      & \B{54.34}   \\ \hline
Llama-3-8B                   & None              &  2:4      & 50.26 & 80.09 & 81.35 & 60.18     & 34.80 & 69.31 & 72.38      & 64.05   \\ \cline{2-11} 
                             & SparseGPT         &  2:4      & 32.00 & 62.67 & 73.70 & 43.19     & 22.20 & 53.79 & 65.75      & 50.47    \\
                             & SparseGPT+SPP     &  2:4      & \B{40.78} & \B{71.09} & 75.35 & 52.01     & 26.40 & 59.93 & \B{67.88}      & 56.21   \\
                             & SparseGPT+SQFT    &  2:4      & 38.05 & 64.02 & 73.27 & 48.89     & 25.20 & 60.65 & 62.12      & 53.17   \\
                             \rowcolor{gray!10}
\cellcolor{white}            & SparseGPT+LoRS    &  2:4      & 40.70 & 70.96 & \B{79.08} & \B{53.26}     & \B{28.00} & \B{60.65} & 67.17      & \B{57.94}    \\ \cline{2-11} 
                             & Wanda             &  2:4      & 26.45 & 55.93 & 66.18 & 37.51     & 18.60 & 52.71 & 60.06      & 45.35    \\
                             & Wanda+SPP         &  2:4      & 38.48 & 68.64 & \B{74.77} & 49.53     & 25.20 & 58.48 & 64.64      & 54.25   \\
                             & Wanda+SQFT        &  2:4      & 37.46 & 65.07 & 73.36 & 49.48     & 26.00 & 63.18 & 62.75      & 53.90   \\ 
                             \rowcolor{gray!10}
\cellcolor{white}            & Wanda+LoRS        &  2:4      & \B{40.78} & \B{70.37} & 77.03 & \B{51.54}     & \B{26.00} & \B{67.87} & \B{64.80}      & \B{56.91}   \\ \hline
\end{tabular}
}
\caption{Zero-shot evaluation results of Llama-2-7b and Llama-3-8b with models trained on the Alpaca dataset. \label{Table: Zero-shot evaluation results from Alpaca}}
\end{table*}

\begin{table*}[t]
\newcolumntype{?}{!{\vrule width 1pt}}
\newcolumntype{C}{>{\centering\arraybackslash}p{2em}}
\centering

\renewcommand\arraystretch{1.5}

\resizebox{\linewidth}{!}{
\begin{tabular}{cllcccccccc}
\hline
Model                        & Method            &  Sparsity & ARC-c & ARC-e & BoolQ & Hellaswag & OBQA  & RTE   & Winogrande & Average \\ \hline
Llama-2-13B                  & None              &  None     & 48.38 & 79.42 & 80.55 & 60.04     & 35.20 & 65.34 & 72.30      & 63.03   \\ \cline{2-11} 
                             & SparseGPT         &  2:4      & 37.29 & 69.07 & \B{79.05} & 48.00     & 25.80 & 58.84 & 69.14      & 55.31   \\
                             & SparseGPT+SPP     &  2:4      & 42.06 & 73.32 & 78.62 & 55.02     & 29.40 & 65.70 & 69.77      & 59.13   \\
                             & SparseGPT+SQFT    &  2:4     & 40.78 & 67.93 & 76.48 & 54.68     & 29.40 & 71.12 & 69.38      & 58.54   \\
                             \rowcolor{gray!10}
\cellcolor{white}            & SparseGPT+LoRS    &  2:4      & \B{42.32} & \B{74.24} & 77.52 & \B{55.81}     & \B{30.00} & 68.95 & \B{70.56}      & \B{59.91}   \\ \cline{2-11} 
                             & Wanda             &  2:4      & 34.47 & 68.48 & 75.72 & 46.39     & 24.40 & 57.04 & 66.69      & 53.31   \\
                             & Wanda+SPP         &  2:4      & 39.42 & 69.40 & 77.37 & 54.84     & \B{30.40} & 65.34 & \B{68.27}      & 57.86   \\
                             & Wanda+SQFT        &  2:4     & 40.02 & 68.35 & 76.09 & 54.17     & 29.80 & 64.98 & 66.93      & 57.19   \\
                             \rowcolor{gray!10}
\cellcolor{white}            & Wanda+LoRS        &  2:4      & \B{41.04} & \B{72.10} & \B{77.46} & \B{55.46}     & 29.40 & \B{68.95} & 67.09      & \B{58.79}   \\ \hline
\end{tabular}
}
\caption{Zero-shot evaluation results of Llama-2-13b trained on the Alpaca dataset.\label{Table: llama-2-13b}}
\end{table*}

\begin{table*}[t]
\newcolumntype{?}{!{\vrule width 1pt}}
\newcolumntype{C}{>{\centering\arraybackslash}p{2em}}
\centering

\renewcommand\arraystretch{1.5}

\resizebox{\linewidth}{!}{
\begin{tabular}{cllcccccccc}
\hline
Model                        & Method            &  Sparsity      & ARC-c & ARC-e & BoolQ & Hellaswag & OBQA  & RTE   & Winogrande & Average \\ \hline
Llama-3-8B                   & SparseGPT         &  unstructured  & 42.66 & 73.95 & 77.16 & 53.86     & 29.40 & 58.84 & 72.30      & 58.31   \\
                             & SparseGPT+SPP     &  unstructured  & 47.53 & \B{77.86} & 80.09 & \B{57.77}     & \B{32.00} & 65.70 & 72.53      & 61.92   \\
                             & SparseGPT+SQFT    &  unstructured  & 46.76 & 77.06 & 80.70 & 56.76     & 30.60 & 64.98 & \B{72.93}      & 61.39   \\
                             \rowcolor{gray!10}
\cellcolor{white}            & SparseGPT+LoRS    &  unstructured  & \B{49.15} & 76.64 & \B{81.71} & 57.66     & 31.00 & \B{69.31} & 72.45      & \B{62.56}   \\ \hline
\end{tabular}
}
\caption{Zero-shot evaluation results of Llama-3-8b trained on the Alpaca dataset under unstructured sparsity.\label{Table: unstructured}}
\end{table*}

\begin{table*}[t]
\newcolumntype{?}{!{\vrule width 1pt}}
\newcolumntype{C}{>{\centering\arraybackslash}p{2em}}
\centering

\renewcommand\arraystretch{1.5}

\resizebox{\linewidth}{!}{
\begin{tabular}{cllcccccccc}
\hline
Model                        & Method            &  Sparsity & ARC-c & ARC-e & BoolQ & Hellaswag & OBQA  & RTE   & Winogrande & Average \\ \hline
Llama-3-8B                   & SparseGPT         &  2:4      & 32.00 & 62.67 & 73.70 & 43.19     & 22.20 & 53.79 & 65.75      & 50.47    \\
                             & SparseGPT+SPP     &  2:4      & \B{39.42} & 69.95 & 71.93 & 51.67     & 25.80 & 63.18 & \B{68.27}      & 55.75   \\
                             & SparseGPT+SQFT    &  2:4      & 38.14 & 70.29 & \B{75.87} & \B{52.35}     & 26.80 & \B{63.90} & 67.56      & 56.42    \\
                             \rowcolor{gray!10}
\cellcolor{white}            & SparseGPT+LoRS    &  2:4      & 39.16 & \B{70.50} & 75.60 & 52.27     & \B{27.40} & 63.54 & 67.48      & \B{56.56}    \\ \cline{2-11} 
                             & Wanda             &  2:4      & 26.45 & 55.93 & 66.18 & 37.51     & 18.60 & 52.71 & 60.06      & 45.35    \\
                             & Wanda+SPP         &  2:4      & 36.77 & 67.39 & \B{72.97} & 49.49     & 25.80 & 59.21 & 64.88      & 53.79   \\
                             & Wanda+SQFT        &  2:4      & 38.31 & 69.53 & 71.56 & \B{50.83}     & \B{28.00} & 54.87 & \B{66.30}      & 54.20   \\ 
                             \rowcolor{gray!10}
\cellcolor{white}            & Wanda+LoRS        &  2:4      & \B{38.57} & \B{69.61} & 72.87 & 50.60     & 27.80 & \B{61.73} & 64.64      & \B{55.12}   \\ \hline
\end{tabular}
}
\caption{Zero-shot evaluation results of Llama-3-8b trained on the SlimPajama dataset with 0.5B tokens.\label{Table: Zero-shot evaluation results from SlimPajama 0.5B}}
\end{table*}

In this section, we aim to demonstrate the efficacy of LoRS in training sparse Large Language Models (LLMs) through a series of experiments.

\vpara{Metrics.} We assessed both the LoRA and SP-LoRA methods based on two primary metrics:

\begin{itemize}[leftmargin=*]
\item  \textbf{Efficiency}: This includes the memory consumption and computational time required during the fine-tuning process.
\item  \textbf{Performance}: We measured this by evaluating the model's accuracy across various downstream tasks.
\end{itemize}


\subsection{Experiment Setup}

Our experimental framework utilized several models from the Llama series: Llama-2-7B, Llama-2-13B and Llama-3-8B~\citep{llama-touvron2023, llama2-touvron2023, llama3-dubey2024}. To create sparse models, we applied post-training pruning techniques, specifically SparseGPT~\cite{sparsegpt-frantar2023} and Wanda~\cite{wanda-sun2024}, using unstructured and 2:4 structured sparsity patterns, following existing works. For the efficiency analysis, we fine-tuned the pruned models with varying batch sizes and adapter ranks to observe their impact on resource utilization. Then, the performance evaluation involved fine-tuning the pruned models on two types of datasets: instruction data and pre-training data. During this phase, adapters were incorporated into all sparse weight matrices within the models.

\begin{itemize}[leftmargin=*]
\item \textbf{Instruction Data}: For instruction tuning, we employed the Stanford-Alpaca dataset \citep{alpaca}. Here, the adapter rank was also set to 16, and the batch size was set to 32 samples, with the learning rate remaining at \(2 \times 10^{-5}\).

\item \textbf{Pre-training Data}: We used a subset of the SlimPajama dataset \citep{refinedweb-penedo2023}, containing 0.5 billion tokens. The setup for this experiment included setting the adapter rank to 16, the batch size to 256,000 tokens, and the learning rate to \(2 \times 10^{-5}\).

\end{itemize}

Following fine-tuning, we evaluated the zero-shot performance of the models on seven benchmark datasets from the EleutherAI LM Evaluation Harness \citep{eval-harness}: ARC-Challenge, ARC-Easy \citep{ARC-clark2018}, BoolQ \citep{boolq-clark2019}, Hellaswag \citep{hellaswag-zellers2019}, OpenBookQA \citep{obqa-mihaylov2018}, RTE, and Winogrande \citep{winogrande-sakaguchi2019}. All experiments were conducted on Nvidia A800-80G GPUs and Nvidia A6000-48G GPUs.

\vpara{Baselines.} To evaluate the effectiveness of LoRS, we compare LoRS with the SP-LoRA methods SPP and SQFT, two existing methods designed to tuning sparse LLMs while preserving sparsity. Refer to Appendix~\ref{Appendix: Conversion of SPP to LoRA variants} for a more detailed explanation.


\subsection{Experiment Results\label{subsection: Memory Overhead}}

\vpara{Efficiency Results.} We evaluated the time and memory overhead of different methods via Llama-3-8B, including LoRS, SQFT and SPP. The implementation details for these methods are provided in Appendix~\ref{Appendix: Implementation}. We conducted experiments for sequence lengths from 512 to 2048 and for adapter ranks from 16 to 64, respectively. Figure~\ref{Figure: time usage} and~\ref{Figure: memory usage} show the time usage and memory usage of the different methods for different sequence lengths, with the adapter rank being 16. It can be seen that LoRS outperforms SPP and SQFT in all scenarios in terms of training throughput and memory overhead, respectively. Compared to SPP, LoRS has a 40\% increase in training speed while having the same memory footprint as SPP. LoRS, on the other hand, saves 40\% of the memory footprint with the same training speed compared to SQFT. Figure~\ref{Figure: time usage rank}, on the other hand, shows how the time overhead varies with the adapter rank size for a sequence length of 2048. It can be seen that changes in the adapter rank size have almost no impact on the time overhead. These results underscore the effectiveness of our approach, demonstrating that LoRS offers an optimal balance between performance and resource utilization when fine-tuning sparse LLMs.

\vpara{Performance Results.} Tables~\ref{Table: Zero-shot evaluation results from Alpaca} and \ref{Table: llama-2-13b} present the zero-shot performance of the Llama-2-7B, Llama-3-8B and Llama-2-13B models, as well as their pruned and fine-tuned variants developed using the Stanford Alpaca under 2:4 sparsity type. Meanwhile, Table~\ref{Table: unstructured} shows the experimental results using unstructured sparsity types. The experimental findings reveal that LoRS significantly boosts the performance of sparse models, with improvements ranging from 7\%\textasciitilde25\% compared to models obtained through post-training pruning. In addition, the effectiveness of LoRS also exceeds that of existing SP-LoRA methods SPP and SQFT. Specifically, LoRS has a 1\%\textasciitilde2\% improvement over SPP and SQFT on the Alpaca dataset due to the better initialization used by LoRS. Table~\ref{Table: Zero-shot evaluation results from SlimPajama 0.5B}, on the other hand, present the results using the SlimPajama-0.5B datasets. On this dataset, LoRS has only minor enhancements compared to SPP and SQFT, this is due to it contains enough data that the impact of initialization on performance is reduced at this point.



\section{Related Work}
\subsection{Pruning}

Pruning is a technique for compressing neural networks by eliminating unimportant weights \citep{pruning-han2016}. It can be divided into structured and unstructured pruning based on the sparsity pattern it induces. Structured pruning removes entire units like channels or layers to simplify the network's architecture. In contrast, unstructured pruning targets individual weights, converting dense matrices into sparse ones. Advances in hardware have enabled efficient execution of models pruned with specific sparse patterns, such as 2:4 sparsity \citep{2:4sparsity-mishra2021}. From an optimization standpoint, pruning methods are also classified as training-based or post-training. Training-based pruning gradually removes weights during the training phase by applying regularization techniques, which can introduce computational overhead and data requirements that are prohibitive for large models \citep{l0-pruning-louizos2018, movementpruning-sanh2020, cofi-xia2022, sp3-hu2024}. Post-training pruning, however, allows for significant model compression using minimal calibration data, making it more suitable for large language models~\citep{sparsegpt-frantar2023, wanda-sun2024, RIA-zhang2024}.

\subsection{Parameter-Efficient Fine-Tuning (PEFT)}

PEFT strategies enable fine-tuning of pre-trained models with minimal parameter updates. These methods typically freeze the original model and introduce trainable adapters, such as prefix tokens, side networks, or parallel/serial adapters \citep{p-tuning-liu2022, sidetuning-zhang2020, adapter-houlsby2019, adapter-family-hu2023}. LoRA and its variants are popular PEFT approaches that allow adapter parameters to merge with model weights after training \citep{lora-hu2021, lora-fa-zhang2023, galore-zhao2024}. However, this merging process can negate the sparsity benefits in sparse LLMs. Our work focuses on adapting LoRA to maintain sparsity.

\subsection{Sparsity-Preserved Training}
Sparsity-preserved training methods aim to train sparse models from the outset or refine existing sparse models. Techniques like STE~\citep{STE-zhou2021}, RigL~\citep{RigL-evci2021}, and others~\citep{AST-huang2024, sparse-finetuning-kurtic2023} ensure that the trained models retain their sparse structure while achieving performance similar to dense counterparts. Despite their potential, these methods often require training all model parameters and can demand more GPU memory than training dense models, presenting challenges for LLM applications. Recent innovations, such as SPP~\cite{spp-lu2024} and SQFT~\cite{sqft-munoz-etal-2024}, attempt to mitigate this issue by integrating PEFT methods with sparsity-preserved training, offering a streamlined approach to training sparse models with reduced costs. Nonetheless, these methods still face high GPU memory overhead due to the construction of full-size matrices during forward passes.
\section{Conclusion}

In this paper, we present LoRS, a novel method designed to train sparse models in a parameter-efficient and memory-efficient manner while preserving sparsity. Our approach specifically tackles the challenges of domain adaptation and performance recovery for sparse large language models (LLMs). By building on the sparsity-preserving LoRA framework, LoRS achieves efficient fine-tuning of LLMs with reduced memory and computation usage through techniques including weight recompute and computational graph reording. Additionally, LoRS enhances the performance of fine-tuned models by employing more effective parameter initialization strategies.

Our experimental results on the Llama family demonstrate that LoRS can efficiently restore the performance of pruned LLMs, surpassing existing methods like SPP and SQFT. This highlights LoRS's potential as an advanced solution for enhancing sparse models without compromising efficiency or performance.

\bibliography{custom}
\appendix

\newpage

\section{Implementation of SPP, SQFT and LoRS \label{Appendix: Implementation}}

\lstset{
    language=Python,          
    basicstyle=\small\ttfamily, 
    keywordstyle=\color{blue},
    commentstyle=\color{red}, 
    stringstyle=\color{magenta},
    showstringspaces=false,   
    breaklines=true,          
    stepnumber=1,             
    numbersep=5pt,            
    backgroundcolor=\color{white}, 
    showspaces=false,         
    showstringspaces=false,   
    showtabs=false,           
    frame=single,             
    tabsize=2,                
    captionpos=b,             
    breaklines=true,          
    breakatwhitespace=false,  
    escapeinside={(*@}{@*)}   
}

\begin{lstlisting}[caption={Implementation of SQFT}]
def forward_sqft(x, W, A, B):
    M = (W != 0)
    W_adapted = W + M * (A @ B)
    return forward_adapter(x, W, A, B)
\end{lstlisting}

\begin{lstlisting}[caption={Implementation of SQFT-gc}]
def forward_ckpt(x, W, A, B):
    M = (W != 0)
    W_adapted = W + M * (A @ B)
    return F.linear(x, W_adapted)

def forward_sqft(x, W, A, B):
    # gradient checkpointing
    return checkpoint(forward_ckpt, x, W, A, B)
\end{lstlisting}

\begin{lstlisting}[caption={Implementation of SPP}]
def forward_adapter(x, W, A, B):
    n, m = W.shape
    r = A.shape[1]
    A = torch.repeat_interleave(weight, m // r, dim=1)
    B = torch.repeat_interleave(weight, n, dim=0)
    W_adapted = W * A * B
    return F.linear(x, W_adapted)

def forward_spp(x, W, A, B):
    y1 = F.linear(x, W)
    y2 = forward_adapter(dropout(x), W, A, B)
    return y1 + y2
\end{lstlisting}

\begin{lstlisting}[caption={Implementation of SPP-gc}]
def forward_adapter(x, W, A, B):
    n, m = W.shape
    r = A.shape[1]
    A = torch.repeat_interleave(weight, m // r, dim=1)
    B = torch.repeat_interleave(weight, n, dim=0)
    W_adapted = W * A * B
    return F.linear(x, W_adapted)

def forward_spp(x, W, A, B):
    y1 = F.linear(x, W)
    # gradient checkpointing
    y2 = checkpoint(forward_adapter, dropout(x), W, A, B)
    return y1 + y2
\end{lstlisting}

\begin{lstlisting}[caption={Implementation LoRS}]
def forward_lors(ctx, x, weight, bias, lora_A, lora_B, params):
    output_shape = x.shape[:-1] + (-1,)
    x_view = x.view(-1, x.shape[-1])
    merged_weight = weight\
        .addmm(lora_A, lora_B, alpha=params.scaling_factor)\
        .mul_(weight != 0)
    y = x_view.mm(merged_weight.t()).view(output_shape)
    y.add_(bias)
    ctx.save_for_backward(x, weight, lora_A, lora_B)
    ctx.params = params
    return y

def backward_lors(ctx, grad_y):
    x, weight, lora_A, lora_B = ctx.saved_tensors
    params = ctx.params
    x_shape = x.shape
    grad_output_shape = grad_y.shape
    x = x.view(-1, x_shape[-1])
    grad_y = grad_y.view(-1, grad_output_shape[-1])
    grad_x = grad_bias = grad_A = grad_B = None
    merged_weight = weight.addmm(lora_A, lora_B, alpha=params.scaling_factor).mul_(weight != 0)
    grad_x = grad_y.mm(merged_weight).view(*x_shape)
    grad_bias = grad_y.sum(dim=0)
    grad_xBt = x @ lora_B.t()
    grad_A = grad_y.t() @ grad_xBt
    grad_yA = grad_y @ lora_A
    grad_B = grad_yA.t() @ x
    return grad_x, None, grad_bias, grad_A, grad_B, None
\end{lstlisting}

\section{SPP and SQFT\label{Appendix: Conversion of SPP to LoRA variants}}




SPP~\citep{spp-lu2024} is a parameter-efficient and sparsity-preserving fine-tuning method. The formulation of SPP can be mathematically described as follows:

\begin{align*}
    & \tilde{\mathcal{W}}^{(t)} = \tilde{\mathcal{W}} + \\
    &\tilde{\mathcal{W}} \odot \text{Repeat}_1(\mathcal{A}^{(t)}, \frac{C}{r}) \odot \text{Repeat}_0(\mathcal{B}^{(t)}, R),    
\end{align*}

where $\tilde{\mathcal{W}} \in \mathbb{R}^{R \times C}$ denotes the sparse weight matrix, $\mathcal{A} \in \mathbb{R}^{R \times r}$ and $\mathcal{B} \in \mathbb{R}^{1 \times C}$ represent the learnable parameter matrices, and $\text{Repeat}_i(x, n)$ means repeating the tensor $x$ along axis $i$ for $n$ times. The adjustment to the weight matrix, denoted by $\tilde{\mathcal{W}} \odot \text{Repeat}_1(\mathcal{A}^{(t)}, \frac{C}{r}) \odot \text{Repeat}_0(\mathcal{B}^{(t)}, R)$, is formulated as the Hadamard product of these three matrices, thereby maintaining the sparsity structure inherent in the matrices involved. Furthermore, the parameters $\mathcal{A}^{(t)}$ and $\mathcal{B}^{(t)}$ are the only ones subject to training, which significantly reduces the parameters compared to that of $\tilde{\mathcal{W}}$, thus exemplifying the parameter efficiency of this approach.






It is observed that SPP can be conceptualized as a variant of LoRA. To illustrate this perspective, consider partitioning each sequence of $r$ consecutive elements within $\mathcal{B}$ into segments, such that:

\begin{equation*}
\mathcal{B} = [\mathcal{B}_1, \mathcal{B}_2, \ldots, \mathcal{B}_{\frac{C}{r}}],
\end{equation*}

where each segment $\mathcal{B}_i$ is a vector of length $r$. Subsequently, we define a block-diagonal matrix $\hat{\mathcal{B}}$ constructed from these segments:

\begin{equation*}
\hat{\mathcal{B}} = [\text{diag}(\mathcal{B}_1), \text{diag}(\mathcal{B}_2), \ldots, \text{diag}(\mathcal{B}_{\frac{C}{r}})].
\end{equation*}

With this definition, the update rule for the weight matrix $\tilde{\mathcal{W}}$ can be rewritten as:

\begin{equation*}
    \tilde{\mathcal{W}}^{(t)} = \tilde{\mathcal{W}} + \tilde{\mathcal{W}} \odot (\mathcal{A}^{(t)} \times \hat{\mathcal{B}}^{(t)}).
\end{equation*}

Therefore, SPP can be interpreted as a LoRA variant that employs a specialized matrix $\hat{\mathcal{B}}$, augmented with the initial weight matrix $\tilde{\mathcal{W}}$ as a weight term, to achieve its parameter-efficient and sparsity-preserving properties.

The distinctions between SPP and LoRA can be delineated as follows:

\begin{itemize}
    \item SPP employs a composite weight matrix $\hat{\mathcal{B}}$ formed by stitching together multiple diagonal matrices, whereas LoRA utilizes a standard matrix $\mathcal{B}$ as its weight matrix.
    \item SPP incorporates the initial weight matrix $\tilde{\mathcal{W}}$ as an additional weight term on the basis of LoRA.
\end{itemize}

SQFT~\citep{sqft-munoz-etal-2024} is another parameter-efficient and sparsity-preserving fine-tuning method. The formulation of SQFT can be mathematically described as follows:

\begin{align*}
    & \tilde{\mathcal{W}}^{(t)} = \tilde{\mathcal{W}} + \mathcal{A}^{(t)} * \mathcal{B}^{(t)} \odot (\tilde{\mathcal{W}} \neq 0)
\end{align*}

where $\tilde{\mathcal{W}} \in \mathbb{R}^{R \times C}$ denotes the sparse weight matrix, $\mathcal{A} \in \mathbb{R}^{R \times r}$ and $\mathcal{B} \in \mathbb{R}^{r \times C}$ represent the learnable parameter matrices.


\end{document}